\documentclass[letterpaper, 10 pt, conference]{ieeeconf}  

\IEEEoverridecommandlockouts                              

\usepackage{amsmath} 
\usepackage{amssymb}  

\usepackage{amsthm}

\usepackage{graphicx} 
\usepackage{subcaption}
\usepackage{multirow}
\usepackage[table]{xcolor}
\usepackage{makecell}
\let\labelindent\relax
\usepackage[shortlabels]{enumitem}
\usepackage{hyperref}
\usepackage{algorithm}
\usepackage{algcompatible}
\usepackage{bm}
\usepackage[export]{adjustbox}

\usepackage{times}

\usepackage{multicol}
\usepackage{amsmath, amssymb}

\usepackage{xspace}
\usepackage{mathtools}
\usepackage[font=small,labelfont=bf,
   justification=justified,
   format=plain]{caption}
\newcommand{\METHOD}{DisCo RL\xspace}
\newcommand{\CMETHOD}{Conditional DisCo RL\xspace}
\newcommand{\METHODLONG}{Distribution-Conditioned Reinforcement Learning\xspace}
\newcommand{\methodlong}{distribution-conditioned reinforcement learning\xspace}

\newcommand{\METHODADJ}{DisCo\xspace}

\newcommand{\gdparam}{\omega}

\newcommand{\gdspace}{\Omega}

\newcommand{\encparam}{{\psi_e}}
\newcommand{\decparam}{{\psi_d}}
\newcommand{\pq}{\phi}
\newcommand{\ppi}{\theta}
\newcommand{\dtask}{\gD_\text{subtask}}

\newcommand{\RS}{RS}
\newcommand{\taskparam}{{\cal{T}}}
\newcommand{\bsigma}{\mathbf{\sigma}}
\newcommand{\blue}[1]{\textcolor{blue}{#1}}

\newcommand{\optional}[1]{#1}

\DeclarePairedDelimiterX{\infdivx}[2]{(}{)}{%
  #1\;\delimsize\|\;#2%
}
\newcommand{\DKL}{D_{\mathrm{KL}}\infdivx*}

\newcommand{\ba}{{\mathbf{a}}}

\newcommand{\bmu}{{\mathbf{\mu}}}

\newcommand{\bs}{{\mathbf s}}
\newcommand{\bS}{{\mathbf S}}
\newcommand{\bz}{{\mathbf z}}

\newtheorem{remark}{Remark}

\usepackage{tikz}
\usepackage{algorithm}
\usepackage{enumitem}
\usepackage{wrapfig}
\usepackage{subcaption}
\usepackage{lipsum}
\usepackage{wrapfig}
\usepackage{pgfplots}

\usepackage{graphicx}
\usepackage{adjustbox}
\usepackage{hyperref}
\usepackage{cleveref}
\setlist{nolistsep}

\usepackage{amsmath,amsfonts,bm}

\def\eqref#1{equation~\ref{#1}}

\def\1{\bm{1}}

\DeclareMathAlphabet{\mathsfit}{\encodingdefault}{\sfdefault}{m}{sl}
\SetMathAlphabet{\mathsfit}{bold}{\encodingdefault}{\sfdefault}{bx}{n}

\def\gA{{\mathcal{A}}}

\def\gC{{\mathcal{C}}}
\def\gD{{\mathcal{D}}}

\def\gM{{\mathcal{M}}}
\def\gN{{\mathcal{N}}}

\def\gR{{\mathcal{R}}}
\def\gS{{\mathcal{S}}}

\def\gZ{{\mathcal{Z}}}

\newcommand{\E}{\mathbb{E}}

\newcommand{\R}{\mathbb{R}}

\DeclareMathOperator*{\argmax}{arg\,max}
\DeclareMathOperator*{\argmin}{arg\,min}

\def\citep{\cite}
\def\citet{\cite}

\title{\LARGE \bf DisCo RL: Distribution-Conditioned Reinforcement Learning for General-Purpose Policies}

\author{Anonymous for Review}
\author{Soroush Nasiriany$^*$, Vitchyr H. Pong$^*$, Ashvin Nair$^*$, Alexander Khazatsky, Glen Berseth, Sergey Levine
\\ 
University of California, Berkeley
\thanks{$^*$Equal contribution. Correspondence: {\tt\small anair17@berkeley.edu}
    }
}

\begin{document}

\maketitle
\begin{abstract}
Can we use reinforcement learning to learn general-purpose policies that can perform a wide range of different tasks, resulting in flexible and reusable skills? Contextual policies provide this capability in principle, but the representation of the context determines the degree of generalization and expressivity. Categorical contexts preclude generalization to entirely new tasks. Goal-conditioned policies may enable some generalization, but cannot capture all tasks that might be desired. In this paper, we propose goal distributions as a general and broadly applicable task representation suitable for contextual policies.
Goal distributions are general in the sense that they can represent any state-based reward function when equipped with an appropriate distribution class, while the particular choice of distribution class allows us to trade off expressivity and learnability.
We develop an off-policy algorithm called \methodlong (\METHOD) to efficiently learn these policies.
We evaluate \METHOD on a variety of robot manipulation tasks and find that it significantly outperforms prior methods on tasks that require generalization to new goal distributions.
\end{abstract}

\section{Introduction}
Versatile, general-purpose robotic systems will require not only broad repertoires of behavioral skills, but also the faculties to quickly acquire new behaviors as demanded by their current situation and the needs of their users.
Reinforcement learning (RL) in principle enables autonomous acquisition of such skills. However, each skill must be learned individually at considerable cost in time and effort.
In this paper, we instead explore how general-purpose robotic policies can be acquired by conditioning policies on task representations.
This question has previously been investigated by learning \emph{goal-conditioned} or \emph{universal} policies, which take in not only the current state, but also some representation of a \emph{goal} state.
However, such a task representation cannot capture many of the behaviors we might actually want a versatile robotic system to perform, since it can only represent behaviors that involve reaching individual states.
For example, for a robot packing items into a box, the task is defined by the position of the items relative to the box, rather than their absolute locations in space, and therefore does not correspond to a single state configuration.
How can we parameterize a more general class of behaviors, so as to make it possible to acquire truly general-purpose policies that, if conditioned appropriately, could perform any desired task?

\begin{figure}[t]
    \centering
    \vspace{-0.05in}
    \includegraphics[width=0.9\linewidth]{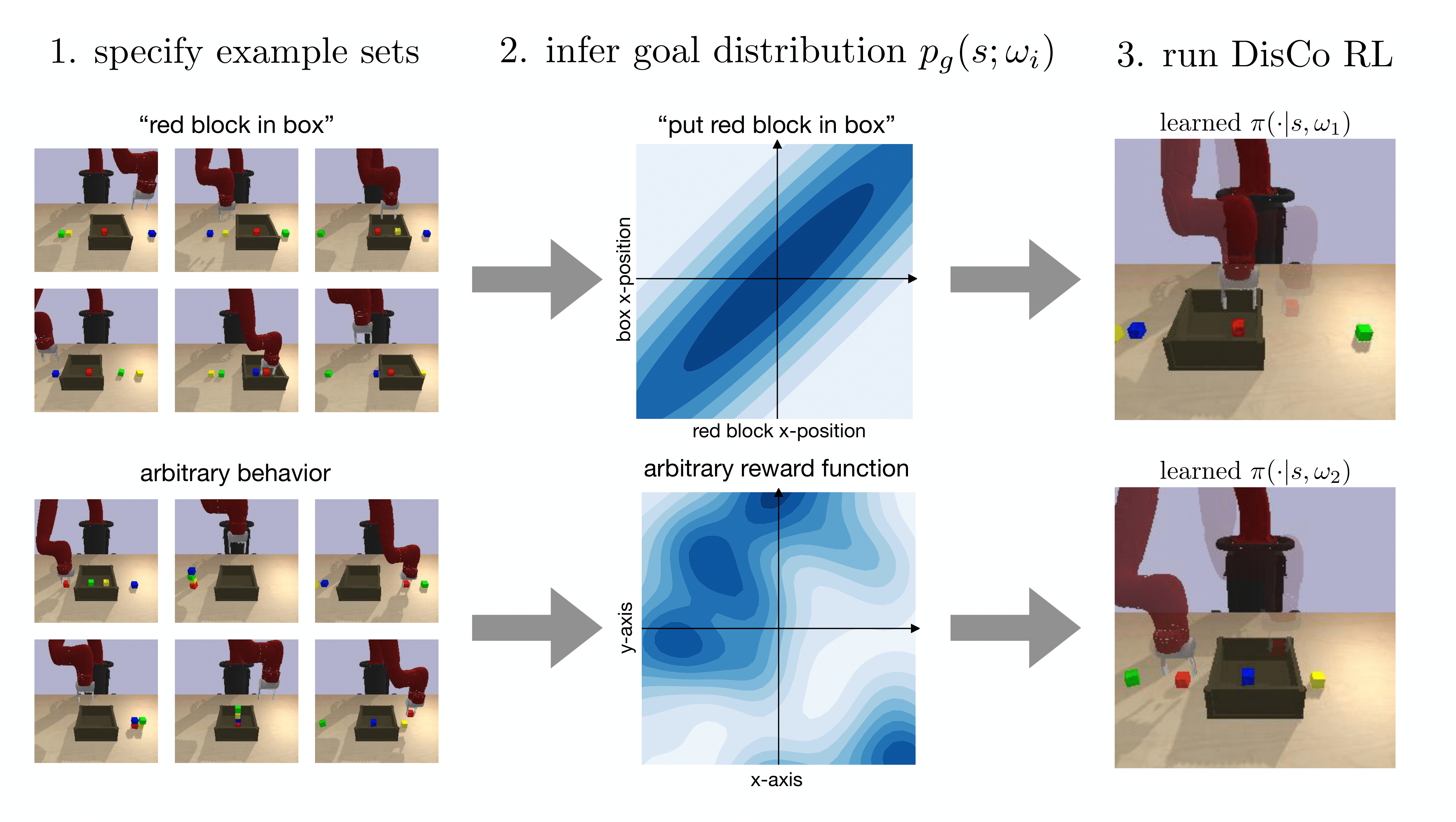}
    \caption{
    We infer the distribution parameters $\gdparam$ from data and pass it to a \METHODADJ policy.
    Distribution-conditioned RL can express a broad range of tasks,
    from defining relationships between different state components (top)
    to more arbitrary behavior (bottom).
    }
    \label{fig:gcdrl-intuition}
    \vspace{-0.50cm}
\end{figure}

To make it possible to learn general-purpose policies that can perform any task, we instead consider conditioning a policy on a full \emph{distribution} over goal states.
Rather than reaching a specific state, a policy must learn to reach states that have high likelihood under the provided distribution, which may specify various covariance relationships (e.g., as shown in Figure~\ref{fig:gcdrl-intuition}, that the position of the items should covary with the position of the box).
In fact, we show that, because optimal policies are invariant to additive factors in reward functions, arbitrary goal distributions can represent \textit{any} state-dependent reward function, and therefore any task.
Choosing a specific distribution class provides a natural mechanism to control the expressivity of the policy.
We may choose a small distribution class to narrow the range of tasks and make learning easier, or we may choose a large distribution class to expand the expressiveness of the policy.

Our experiments demonstrate that distribution-conditioned policies can be trained efficiently by sharing data from a variety of tasks and relabeling the goal distribution parameters, where each distribution corresponds to a different task reward.
Lastly, while the distribution parameters can be provided manually to specify tasks, we also present two ways to infer these distribution parameters directly from data.

The main contribution of this paper is \METHOD, an algorithm for learning distribution-conditioned policies.
To learn efficiently, \METHOD uses off-policy training and a novel distribution relabeling scheme.
We evaluate on robot manipulation tasks in which a single policy must solve multiple tasks that cannot be expressed as reaching different goal states.
We find that conditioning the policies on goal distributions results in significantly faster learning than solving each task individually, enabling policies to acquire a broader range of tasks than goal-conditioned methods.

\section{Related Work}
In goal-conditioned RL, a policy is given a goal state, and must take actions to reach that state~\citep{kaelbling1993goals,schaul2015uva,pathak2018zeroshot,dhiman2018floyd,pong2018tdm,nachum2018hiro,nair2018rig,wardefarley2019discern,pong2019skewfit,jurgenson2019sub}.
However, as discussed previously, many tasks cannot be specified with a single goal state.
To address this, many goal-conditioned methods manually design a goal space that explicitly excludes some state variables~\citep{andrychowicz2017her,levy2017hierarchical,rauber2017hindsight,florensa2018automatic,plappert2018multigoal,colas2019curious}, for example by only specifying the desired location of an object.
This requires manual effort and user insight, and does not generalize to environments with high-dimensional state representations, such as images, where manually specifying a goal space is very difficult.
With images, a number of methods learn latent representations for specifying goal states~\citep{laversanne2018curiosity,nair2018rig,pong2019skewfit,nair2020contextual,nair2020goalaware}, which makes image-based goals more tractable, but does not address the representation issues discussed above.
Our work on goal distributions can be seen as a generalization of goal state reaching.
Reaching a single goal state $g$ is equivalent to maximizing the likelihood of a delta-distribution centered at $g$, and
ignoring some state dimensions is equivalent to maximizing the likelihood of a distribution that places uniform likelihood across the respective ignored state variables.
But more generally, goal distributions capture the set of all reward functions, enabling policies to be conditioned on arbitrary tasks.

A number of prior methods learn rewards~\citep{torabi2018generative,fu2018variational} or policies~\cite{torabi2018behavioral,sermanet2018time,liu2018imitation,edwards2019imitating} using expert trajectories or observations.
In this work, we also demonstrate that we can use observations to learn reward functions, but we have different objectives and assumptions as compared to prior work.
Many of these prior methods 
require state sequences from expert demonstrations~\citep{torabi2018behavioral,sermanet2018time,liu2018imitation,edwards2019imitating},
whereas our work only requires observations of successful outcomes to fit the goal distribution.
\citet{fu2018variational} also only uses observations of successful outcomes to construct a reward function, but focuses on solving single tasks or goal-reaching tasks, whereas we study the more general setting where the policy is conditioned on a goal distribution.
Parametric representations of rewards have also been in the context of successor features~\citep{kulkarni2016deep,barreto2017successor,borsa2018universal,barreto2018transfer},
which parameterize reward functions as linear combinations of known features.
We present a general framework in which arbitrary rewards can be represented as goal distributions rather than feature weights, and also demonstrate that these goal distributions can be learned from data.

Prior work on state marginal matching~\citep{lee2019efficient} attempts to make a policy's stationary distribution match a target distribution to explore an environment.
In our work, rather than matching a target distribution, we use the log-likelihood of a goal distributions to define a reward function, which we then maximize with standard reinforcement learning.

\section{Background}\label{sec:background}
Reinforcement learning (RL) 
frames reward maximization in a Markov decision process (MDP), defined by the tuple $\gM = (\gS, \gA, r, p, p_{0}, \gamma )$~\citep{sutton1998rl}, where $\gS$ denotes the state space and $\mathcal{A}$ denotes the action space.
In each episode, the agent's initial state $\bs_0 \in \gS$ is
sampled from an initial state distribution $\bs_0 \sim p_0(\bs_0)$, the agent chooses an action $\ba \in \mathcal{A}$ according to a stochastic policy $\ba_t \sim \pi( \cdot | \bs_t)$, and the next state is generated from the state transition dynamics $\bs_{t+1} \sim p(\cdot \mid \bs_t, \ba_t)$.
We will use $\tau$ to denote a trajectory sequence $(\bs_0, \ba_0, \bs_1, \dots)$ and denote sampling as \mbox{$\tau \sim \pi$} since we assume a fixed initial state and dynamics distribution.
The objective of an agent is to maximize the sum of discounted rewards,
    $\E_{\tau \sim \pi}\left[\sum_{t=1}^\infty \gamma^t r(\bs_t, \ba_t)\right]$.

\noindent \textbf{Off-policy, temporal-difference algorithms.}
Our method can be used with any off-policy temporal-difference (TD) learning algorithm.
TD-learning algorithms only need $(\bs_{t}, \ba_{t}, r_{t}, \bs_{t+1})$ tuples to train a policy, where $\ba_{t}$ is an action taken from state $\bs_{t}$, and where $r_{t}$ and $\bs_{t+1}$ are the resulting reward and next state, respectively, sampled from the environment.
Importantly, these tuples can be collected by any policy, making it an off-policy algorithm.

A contextual MDP augments an MDP with a context space $\gC$, and both the reward and the policy \footnote{Some work assumes the dynamics $p$ depend on the context~\citep{hallak2015contextual}. We assume that the dynamics are the same across all contexts, though our method could also be used with context-dependent dynamics.} are conditioned on a context $c \in \gC$ for the entire episode. 
Contextual MDPs allow us to easily model policies that accomplish much different tasks, or, equivalently, maximize different rewards, at each episode.
The design of the context space and contextual reward function determines the behaviors that these policies can learn.
If we want to train general-purpose policies that can perform arbitrary tasks, is it possible to design a contextual MDP that captures the set of all rewards functions?
In theory, the context space $\gC$ could be the set of all reward functions.
But in practice, it is unclear how to condition policies on reward functions, since most learning algorithms are not well suited to take functions as inputs.
In the next section, we present a promising context space: the space of goal distributions.
\section{\METHODLONG}
In this section, we show how conditioning policies on a goal distribution results in a contextual MDP that can capture any set of reward functions.
Each distribution represents a different reward function, and so choosing a distribution class provides a natural mechanism to choose the expressivity of the contextual policy.
We then present \methodlong (\METHOD), an off-policy algorithm for training policies conditioned on parametric representation of distributions, and discuss the specific representation that we use.

\subsection{Generality of Goal Distributions}
We assume that the goal distribution is in a parametric family, with parameter space $\gdspace$, and consider a contextual MDP in which the context space is $\gdspace$.
At the beginning of each episode, a parameter $\gdparam \in \gdspace$ is sampled from some parameter distribution $p_\gdparam$.
The parameter $\gdparam$ defines the goal distribution \mbox{$p_g(\bs ; \gdparam): \gS \mapsto \R_+$} over the state space.
The policy is conditioned on this parameter, and is given by $\pi(\cdot \mid \bs, \gdparam)$.
The objective of a distribution-conditioned (\METHODADJ) policy is to reach states that have high log-likelihood under the goal distribution, which can be expressed as
\begin{align}\label{eq:gdc-objective}
    \max_\pi \E_{\tau \sim \pi(\cdot \mid \bs, \gdparam)}\left[\sum_t \gamma^t \log p_g(\bs_t; \gdparam)\right].
\end{align}
This formulation can express arbitrarily complex distributions and tasks, as we illustrate in \autoref{fig:gcdrl-intuition}. More formally:
\begin{remark}
Any reward maximization problem
can be equivalently written as maximizing the log-likelihood under a goal distribution (\autoref{eq:gdc-objective}), up to a constant factor.
\end{remark}
This statement is true because, for any reward function of the form $r(\bs)$, we can define a distribution \mbox{$p_g(\bs) \propto e^{r(\bs)}$},
from which we can conclude that maximizing $\log p_g(\bs)$ is equivalent to maximizing $r(\bs)$, up to the constant normalizing factor in the denominator.
If the reward function depends on the action, $r(\bs, \ba)$, we can modify the MDP and append the previous action to the state $\bar{\bs} = [\bs, \ba]$, reducing it to another MDP with a reward function of the form $r(\bar{\bs})$.

Of course, while any reward can be expressed as the log-likelihood of a goal distribution, a specific fixed parameterization $p_g(\bs ; \gdparam)$ may not by itself be able to express any reward.
In other words, choosing the distribution parameterization is equivalent to choosing the set of reward functions that the conditional policy can maximize.
As we discuss in the next section, we can trade-off expressivity and ease of learning by choosing an appropriate goal distribution family.

\subsection{Goal Distribution Parameterization}\label{sec:goal-distribution-parameterization}

Different distribution classes represent different types of reward functions.
To explore the different capabilities afforded by different distributions, we study three families of distributions.
\paragraph{Gaussian distribution}
A simple class of distributions is the family of multivariate Gaussian.
Given a state space in $\R^n$, the distribution parameters consists of two components, \mbox{$\gdparam = (\bmu, \Sigma)$}, where \mbox{$\bmu \in \R^n$} is the mean vector and \mbox{$\Sigma \in \R^{n \times n}$} is the covariance matrix.
When inferring the distribution parameters with data, we regularize the $\Sigma^{-1}$ matrix by thresholding absolute values below $\epsilon = 0.25$ to zero.
With these parameters, the reward from \autoref{eq:gdc-objective} is given by
\begin{align}\label{eq:gaussian-log-prob-reward}
    r(\bs; \omega) = -0.5 (\bs - \bmu)^T \Sigma^{-1} (\bs - \bmu),
\end{align}
where we have dropped constant terms that do not depend on the state $\bs$.
Using this parameterization, the weight of individual state dimensions depend on the values along the diagonal of the covariance matrix.
By using off-diagonal covariance values, this parameterization also captures the set of tasks in which state components need to covary, such as when the one object must be placed near another one, regardless of the exact location of those objects (see the top half of \autoref{fig:gcdrl-intuition}).

\paragraph{Gaussian mixture model}
A more expressive class of distributions that we study is the Gaussian mixture model with 4 modes, which can represent multi-modal tasks.
\optional{
The parameters are the mean and covariance of each Gaussian and the weight assigned to each Gaussian distribution. The reward is given by the log-likelihood of a state.
}

\paragraph{Latent variable model}
To study even more expressive distributions, we consider a class of distributions parameterized by neural networks.
These models can be extremely expressive~\citep{dinh2016density, van2016conditional}, often at the expense of having millions of parameters and requiring large amounts of training data~\citep{dinh2016density, kingma2018glow}.

To obtain an expressive yet compact parameterization, we consider non-linear reparameterizations of the original state space.
Specifically, we model a distribution over the state space using a Gaussian variational auto-encoder (VAE)~\citep{kingma2014vae,rezende2014stochasticbackprop}.
Gaussian VAEs model a set of observations using a latent-variable model of the form
\begin{align*}
    p(\bs) = \int_{\gZ} p(\bz) p_\decparam(\bs \mid \bz) d\bz,
\end{align*}
where $\bz \in \gZ = \mathbb{R}^{d_z}$ are latent variables with dimension $d_z$.
The distribution $p_\decparam$ is a learned generative model or ``decoder'' and $p(z)$ is a standard multivariate Gaussian distribution in $\R^{d_z}$.
A Gaussian VAE also learns a posterior distribution or ``encoder'' that maps states $\bs$ onto Gaussian distributions in a latent space, given by $q_\encparam(\bz; \bs)$.

Gaussian VAEs are explicitly trained so that Gaussian distributions in a latent space define distributions over the state space.
Therefore, we represent a distribution over the state space with the mean $\bmu_z \in \mathbb{R}^{d_z}$ and variances $\bsigma_z \in \mathbb{R}^{d_z}$ of a diagonal Gaussian distribution \emph{in the learned, latent space}, which we write as $\gN(\bz; \bmu_z, \bsigma_z)$.
In other words, the parameters $\gdparam = (\bmu_z, \bsigma_z )$ define the following distribution over the states:
\begin{align}\label{eq:non-linear-param}
    p_g(\bs; \omega) = \int_{\gZ} \gN (\bz ; \bmu_z, \bsigma_z) p_\decparam(\bs \mid \bz) d\bz,
\end{align}
where $p_\decparam(\bs \mid \bz)$ is the generative model from the VAE.
Because $\bz$ resides in a learned latent space, the set of reward functions that can be expressed with \Cref{eq:non-linear-param} includes arbitrary non-linear transformations of $\bs$.

\section{Learning Goal Distributions and Policies}\label{sec:goal-distribution-inference}
Given any of the distribution classes mentioned above, we now consider how to obtain a specific goal distribution parameter $\gdparam$ and train a policy to maximize \Cref{eq:gdc-objective}.
We start with obtaining goal distribution parameters $\gdparam$.
While a user can manually select the goal distribution parameters $\gdparam$, this requires a degree of user insight, which can be costly or practically impossible if using the latent representation in \Cref{eq:non-linear-param}.
We discuss two automated alternatives for obtaining the goal distribution parameters $\gdparam$.

\subsection{Inferring Distributions from Examples}
\label{sec:infer-from-examples}
One simple and practical way to specify a goal distribution is to provide $K$ example observations $\{\bs_k\}_{k=1}^K$ in which the task is successfully completed.
This supervision can be easier to provide than full demonstrations, which not only specify the task but also must show how to solve the task through a sequence of states $(\bs_1, \bs_2, \dots)$ or states and actions $(\bs_1, \ba_1, \bs_2, \dots)$.
Given the example observations, we describe a way to infer the goal parameters based on the parameterization.

If $\gdparam$ represents the parameters of a distribution in the state space, we learn a goal distribution via maximum likelihood estimation (MLE), as in
\begin{align}\label{eq:state-gd-inference}
    \gdparam^* = \argmax_{\gdparam \in \gdspace} \sum_{k=1}^K \log p_g(\bs_k; \gdparam),
\end{align}
and condition the policy on the resulting parameter $\gdparam^*$.
If $\gdparam$ represents the parameters of a distribution in the latent space, we need a Gaussian distribution in the latent space that places high likelihood on all of the states in $\dtask$.
We obtain such a distribution by finding a latent distribution that minimizes the KL divergence to the mixture of posteriors $\frac{1}{K} \sum_k q_\encparam(\gdparam ; \bs_k)$.
Specifically, we solve the problem
\begin{align}\label{eq:latent-space-gd-inference}
    \gdparam^* = \argmin_{\gdparam \in \gdspace} \DKL{
    \frac{1}{K} \sum_k q_\encparam(\bz ; \bs_k)
    }{p(\bz; \gdparam)},
\end{align}
where $\gdspace$ is the set of all means and diagonal covariance matrices in $\R^{d_z}$.
The solution to \Cref{eq:latent-space-gd-inference} can be computed in closed form using moment matching~\citep{minka2001expectation}.

\begin{figure}[t]
    \centering
    \includegraphics[width=\linewidth]{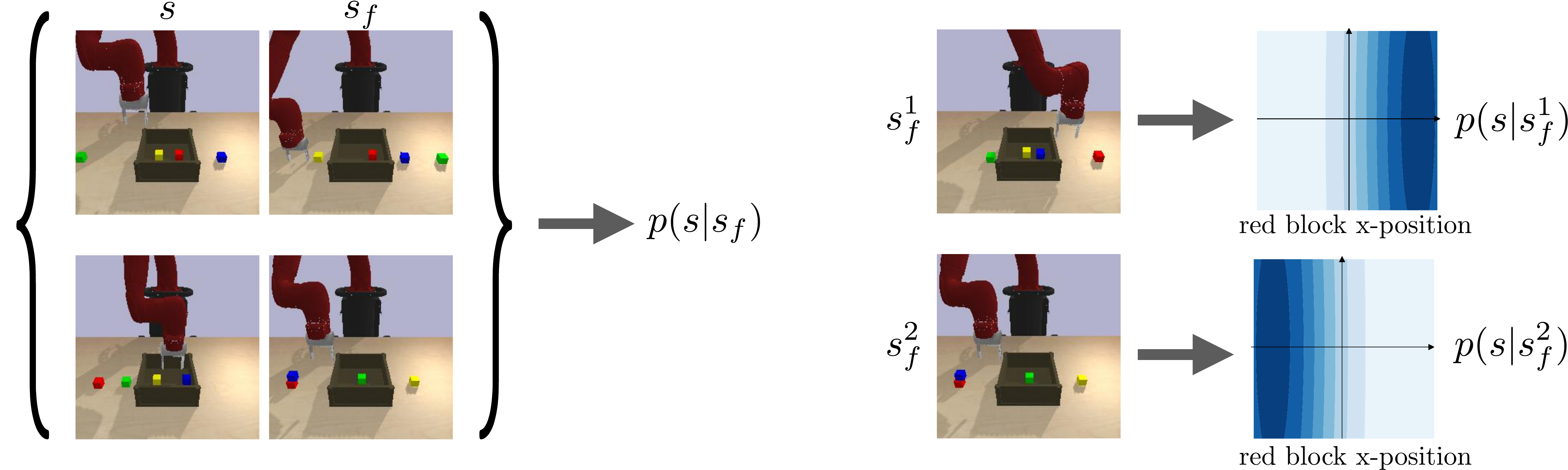}
    \caption{(Left) A robot must arrange objects into a configuration, $\bs_f$, that varies between episodes.
    This task contains sub-tasks, such as first moving the red object to the correct location.
    Given a final state $\bs_f$, there exists a distribution of intermediate states $\bs$ in which this first sub-task is completed.
    We use pairs of states $(\bs, \bs_f)$ to learn a conditional distribution \mbox{$p(\bs \mid \bs_f)$}, which (right) defines the first sub-task given the final state, $\bs_f$.}
    \label{fig:conditional-distribution-visualization}
    \vspace{-0.5cm}
\end{figure}

\subsection{Dynamically Generating Distributions via Conditioning}
\label{sec:infer-conditional}

We also study a different use case for training a \METHODADJ policy: automatically decomposing long-horizon tasks into sub-tasks.
Many complex tasks lend themselves to such a decomposition. For example, a robot that must arrange a table can divide this task into placing one object at a desired location before moving on to the next object.
Through this decomposition, the agent can learn to solve the task significantly faster than solving the entire task directly.
To operationalize this idea, we need a representation for sub-tasks and a method to automatically decompose a task into a set of sub-tasks.
Moreover, we would like to train a single policy that can accomplish all of these sub-tasks.
How can we satisfy all of these objectives?

One potential approach is to represent each sub-task as a single goal state and train a single goal-conditioned policy that satisfies these sub-tasks.
However, this representation is overly restrictive, as most sub-tasks can be satisfied by a set of goal states rather than a single goal state.
This representation therefore restricts the class of possible sub-tasks and makes the problem artificially harder. We can address these limitations by training a \METHODADJ policy, which represents each sub-task as a goal distribution.
Specifically, a task represented by parameters $\taskparam$ can be decomposed into a sequence of $M$ sub-tasks, each of which in turn is represented by a goal distribution.
We accomplish this by learning \textit{conditional distributions}, $p_g^i(\bs \mid \taskparam)$: conditioned on some final task $\taskparam$, the conditional distribution $p_g^i(\bs \mid \taskparam)$ is a distribution over desired states for sub-task i.

To obtain a conditional distribution $p_g^i(\bs \mid \taskparam)$ for sub-task $i$, we assume access to tuples \mbox{$\dtask ^i = \{(\bs^{(k)}, \taskparam^{(k)})\}_{k=1}^K$}, where $\bs^{(k)}$ is an example state in which sub-task $i$ is accomplished when trying to accomplish a task $\taskparam^{(k)}$.
See \Cref{fig:conditional-distribution-visualization} for a visualization.
In our experiments, we study the case where final tasks are represented by a final state $\bs_f$ that we want the robot to reach, meaning that \mbox{$\taskparam = \bs_f$}.
We note that one can train a goal-conditioned policy to reach this final state $\bs_f$ directly, but, as we will discuss in \Cref{sec:experiments}, this decomposition significantly accelerates learning by exploiting access to the pairs of states.

For each $i$, and dropping the dependence on $i$ for clarity, we learn a conditional distribution, by fitting a joint Gaussian distribution, denoted by \mbox{$p_{\bS, \bS_f}(\bs, \bs_f; \bmu, \Sigma)$}, to these pairs of states using MLE, as in
\begin{align}\label{eq:learn-joint-distrib}
    \bmu^*, \Sigma^* = \argmax_{\mu, \Sigma} \sum_{k=1}^K \log p_{\bS, \bS_f}(\bs^{(k)}, \bs_f^{(k)}; \bmu, \Sigma).
\end{align}
This procedure requires up-front supervision for each sub-task during training time, but at test time, given a desired final state $\bs_f$, 
we compute the parameters of the Gaussian conditional distribution $p_g(\bs \mid \bs_f; \mu^*, \Sigma^*)$ in closed form. Specifically, the conditional parameters are given by
\begin{equation}
  \label{eq:conditional-params}
  \begin{aligned}
\bar \mu &= \mu_1 + \Sigma_{12} \Sigma_{22}^{-1}(\bs_f - \mu_2)\\
\bar \Sigma &= \Sigma_{11} - \Sigma_{12} \Sigma_{22}^{-1} \Sigma_{21},
  \end{aligned}
\end{equation}
where $\mu_1$ and $\mu_2$ represents the first and second half of $\mu^*$, and similarly for the covariance terms.
To summarize, we learn $\mu^*$ and $\Sigma^*$ with \Cref{eq:learn-joint-distrib} and then use \Cref{eq:conditional-params} to automatically transform a final task $\taskparam = \bs_f$ into a goal distribution $\gdparam = (\bar \mu, \bar \Sigma)$ which we give to a \METHODADJ policy.

Having presented two ways to obtain distribution parameters, we now turn to learning \METHOD policies.

\subsection{Learning Distribution-Conditioned Policies}\label{sec:learning-disco-policies}
In this section, we discuss how to optimize \autoref{eq:gdc-objective} using an off-policy TD algorithm.
As discussed in \autoref{sec:background}, TD algorithms require tuples of state, action, reward, and next state, denoted by $(\bs_{t}, \ba_{t}, r_{t}, \bs_{t+1})$.
To collect this data, we condition a policy on a goal distribution parameter $\gdparam$, collect a trajectory with the policy \mbox{$\tau = [\bs_0, \ba_0, \cdots]$}, and then store the trajectory and distribution parameter into a replay buffer~\citep{mnih2015human}, denoted as $\gR$.
We then sample data from this replay buffer to train our policy using an off-policy TD algorithm.
We use soft actor-critic as our RL algorithm~\citep{haarnoja2018sac}, though in theory any off-policy algorithm could be used.
Because TD algorithms are off-policy, we propose to reuse data collected by a policy conditioned on one goal distribution $\gdparam$ to learn about how a policy should behave under another goal distribution $\gdparam'$.
In particular, given a state $\bs$ sampled from a policy that was conditioned on some goal distribution parameters $\gdparam$, we occasionally relabel the goal distribution with an alternative goal distribution \mbox{$\gdparam' = \RS(\bs, \gdparam)$} for training, where $\RS$ is some relabeling strategy.
For relabeling, we would like to create a distribution parameter that gives high reward to an achieved state.
We used a simple strategy that we found worked well:
we replace the mean with the state vector $\bs$ and randomly re-sample the covariance from the set of observed covariances as in $\RS(\bs, (\mu, \Sigma)) =  (\bs, \Sigma')$, where $\Sigma'$ is sampled uniformly from the replay buffer.
This relabeling is similar to relabeling methods used in goal-conditioned reinforcement learning~\citep{kaelbling1993goals,andrychowicz2017her,rauber2017hindsight,pong2018tdm,nachum2018hiro,plappert2018multigoal,nair2018rig,eysenbach2020rewriting}.

\vspace{0.3cm}
\begin{algorithm}[t]
\caption{\blue{(Conditional)} Distribution-Conditioned RL}
\label{algo:disco}
\begin{algorithmic}[1]
\REQUIRE Policy $\pi_\ppi$, $Q$-function $Q_\pq$, TD algorithm $\gA$, relabeling strategy $\RS$, exploration parameter distribution $p_\gdparam$, replay buffer $\gR$.
\STATE Compute $\gdparam$ using Equation (4) or (5) \blue{(if unconditional)}.
\FOR{$0, \dots, N_\text{episode}-1$ episodes}
    \STATE \blue{Sample $\bs_f$ and compute $\gdparam$ with \Cref{eq:conditional-params}.}
    \STATE Sample trajectory from environment $\tau \sim \pi(\cdot | \bs; \gdparam)$.
    \STATE Store tuple $(\tau, \gdparam)$ in replay buffer $\gR$.
    \FOR{$0, \dots,  N_\text{updates}-1$ steps}
        \STATE Sample trajectory and parameter $(\tau, \gdparam) \sim \gR$.
        \STATE Sample transition tuple $(\bs_t, \ba_t, \bs_{t+1})$ and a future state $\bs_h$ from $\tau$, where $t < h$.
        \STATE With probability $p_\text{relabel}$, relabel $\gdparam \leftarrow \RS(\bs_h, \gdparam)$.
        \STATE Compute reward $r = \log p_g(\bs_t ; \gdparam)$ and augment states $\hat \bs_t \leftarrow [\bs_t; \gdparam]$, $\hat \bs_{t+1} \leftarrow [\bs_{t+1}; \gdparam]$.
        \STATE Update $Q_\pq$ and $\pi_\ppi$ using $\gA$ and $(\hat \bs_t; \ba_t, \hat \bs_{t+1}, r)$.
    \ENDFOR
\ENDFOR
\end{algorithmic}
\end{algorithm}
We call the overall algorithm \textit{\METHOD} when training a distribution from examples and call the variant that learns a conditional distribution \textit{\CMETHOD}.
We present a summary of the method in \Cref{algo:disco}.
\section{Experiments}\label{sec:experiments}

Our experiments study the following questions:
\textbf{(1)} How does \METHOD with a learned distribution compare to prior work that also uses successful states for computing rewards?
\textbf{(2)} Can we apply \CMETHOD to solve long-horizon tasks that are decomposed into shorter sub-tasks?
\textbf{(3)} How do \METHODADJ policies perform when conditioned on goal distributions that were never used for data-collection?
The first two questions evaluate the variants of \METHOD presented in \Cref{sec:goal-distribution-inference}, and the third question studies how well the method can generalize to test time task specifications.
We also include ablations that study the importance of the relabeling strategy presented in \autoref{sec:learning-disco-policies}.
Videos of our method and baselines are available on the paper website.\footnote{\url{https://sites.google.com/view/disco-rl}}

\begin{figure}[t]
    \begin{subfigure}{0.33\linewidth}
        \centering
        \includegraphics[width=0.85\textwidth]{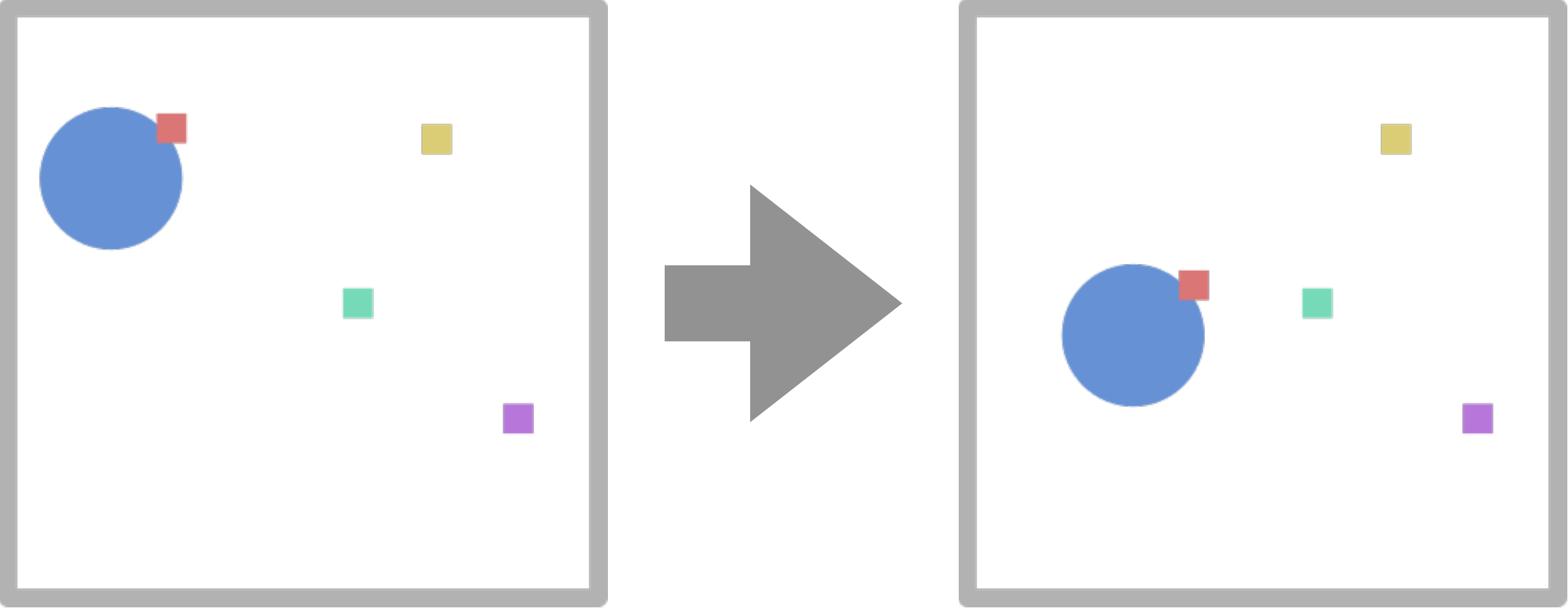}\\
    \vspace{0.1cm}
        \includegraphics[width=0.85\textwidth]{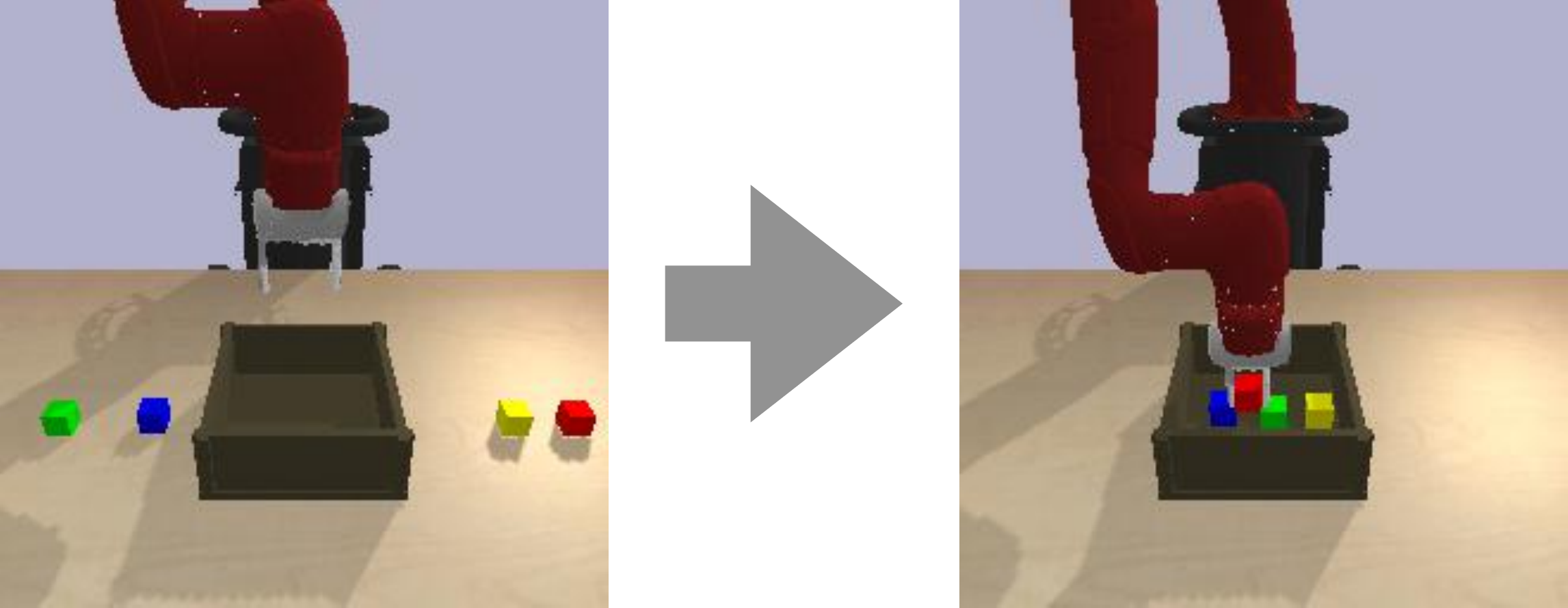}
    \end{subfigure}
    \begin{subfigure}{0.66\linewidth}
        \centering
        \includegraphics[height=2.0cm]{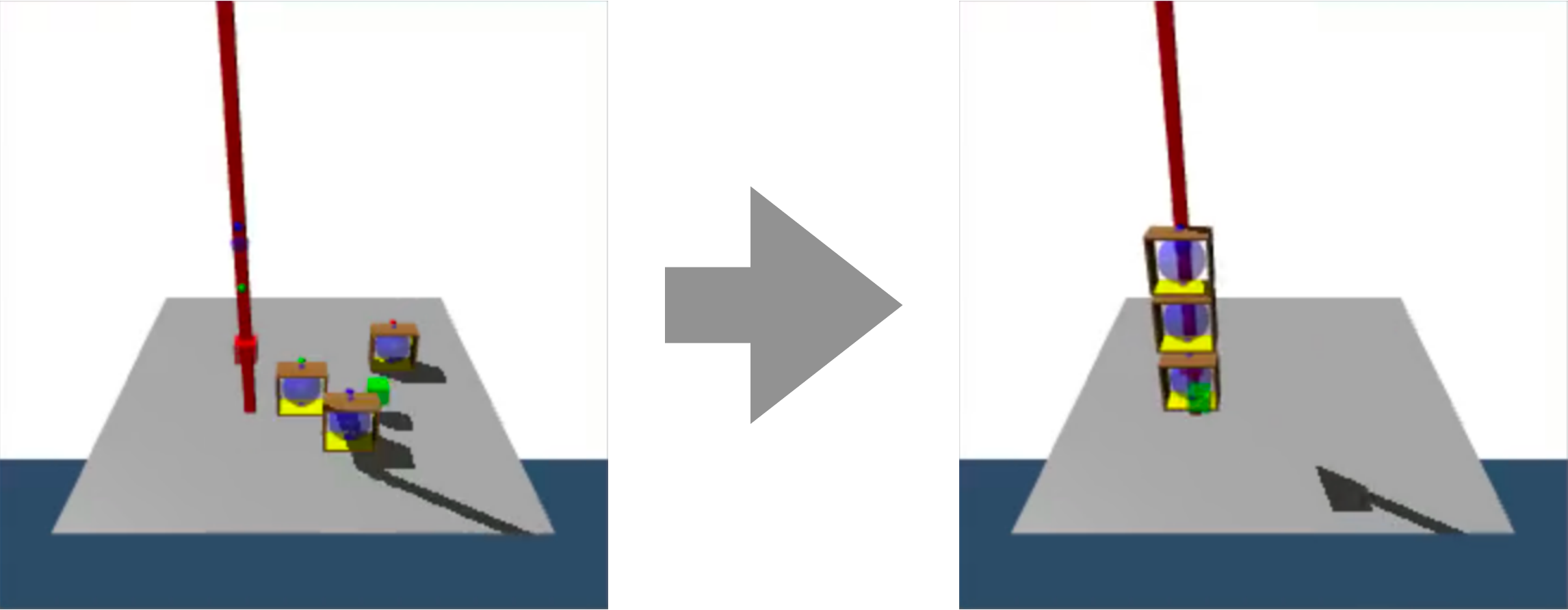}
    \end{subfigure}
    \caption{Illustrations of the experimental domains, in which a policy must
    (left top) use the blue cursor to move objects to different locations,
    (left bottom) control a Sawyer arm to move cubes into and out of a box,
    and
    (right) attach shelves to a pole using a cursor.
    }
    \label{fig:env-pictures}
    \vspace{-0.3cm}
\end{figure}

We study these questions in three simulated manipulation environments of varying complexity, shown in \autoref{fig:env-pictures}.
We first consider a simple two-dimensional ``Flat World'' environment in which an agent can pick up and place objects at various locations.
The second environment contains a Sawyer robot, a rectangular box, and four blocks, which the robot must learn to manipulate.
The agent controls the velocity of the end effector and gripper, and the arm is restricted to move in a 2D plane perpendicular to the table's surface.
Lastly, we use an IKEA furniture assembly environment from \citet{lee2019ikea}.
An agent controls the velocity of a cursor that can lift and place $3$ shelves onto a pole.
Shelves are connected automatically when they are within a certain distance of the cursor or pole.
The states comprise the Cartesian position of all relevant objects and, for the Sawyer task, the gripper state.
For the Sawyer environment, we also consider an image-based version which uses a 48x48 RGB image for the state.
To accelerate learning on this image environment, we follow past work on image-based RL~\citep{nair2018rig} and encode images $s$ with the mean of the VAE encoder $q_\encparam(\bz; \bs)$ for the RL algorithm and pretrain the VAE on 5000 randomly collected images.
All plots show mean and standard deviation across 5 seeds, as well as optimal performance with a dashed line.

\subsection{Learning from Examples}
Our first set of experiments evaluates how well \METHOD performs when learning distribution parameters from a fixed set of examples, as described in \Cref{sec:infer-from-examples}.
We study the generality of \METHOD across all three parameterizations from \Cref{sec:goal-distribution-parameterization} on different environments:
First, we evaluate \METHOD with a Gaussian model learned via \Cref{eq:state-gd-inference} on the Sawyer environment, in which the policy must move the red object into the box while ignoring the remaining three ``distractor'' objects.
Second, we evaluate \METHOD with a GMM model also learned via \Cref{eq:state-gd-inference} on the Flat World environment, where the agent must move a specific object to any one of four different locations.
Lastly, we evaluate \METHOD with a latent variable model learned via \Cref{eq:latent-space-gd-inference} on an image-based version of the Sawyer environment, where the policy must move the hand to a fixed location and ignore visual distractions.
We note that this last experiment is done completely from images, and so manually specifying the parameters of a Gaussian in image-space would be impractical.
The experiments used between $K=30$ to $50$ examples each for learning the goal distribution parameters.
We report the normalized final distance: a value $1$ is no better than a random policy.

We compare to past work that uses example states to learn a reward function.
Specifically, we compare to variational inverse control with events (\textbf{VICE})~\citep{fu2018variational}, which trains a success classifier to predict the user-provided example states as positive and replay buffer states as negative, and then uses the log-likelihood of the classifier as a reward.
We also include an oracle labeled \textbf{SAC (oracle reward)} which uses the ground truth reward.
Since VICE requires training a separate classifier for each task, these experiments only test these methods on a single task.
Note that a single \METHOD policy can solve multiple tasks by conditioning on different parameter distributions, as we will study in the next section.

\begin{figure}
    \centering
    
    \includegraphics[height=2.2cm]{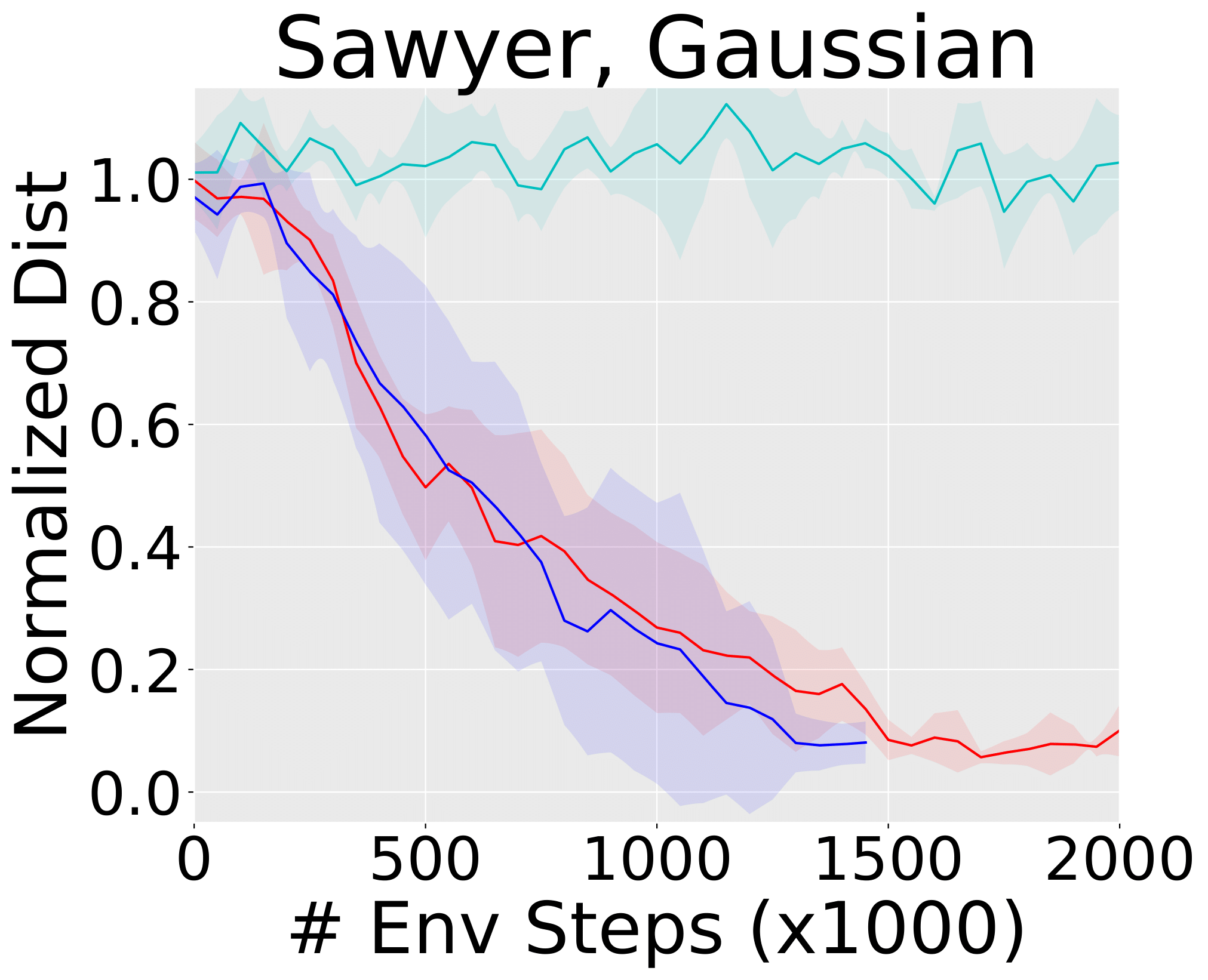}
    \hfill
    \includegraphics[height=2.2cm]{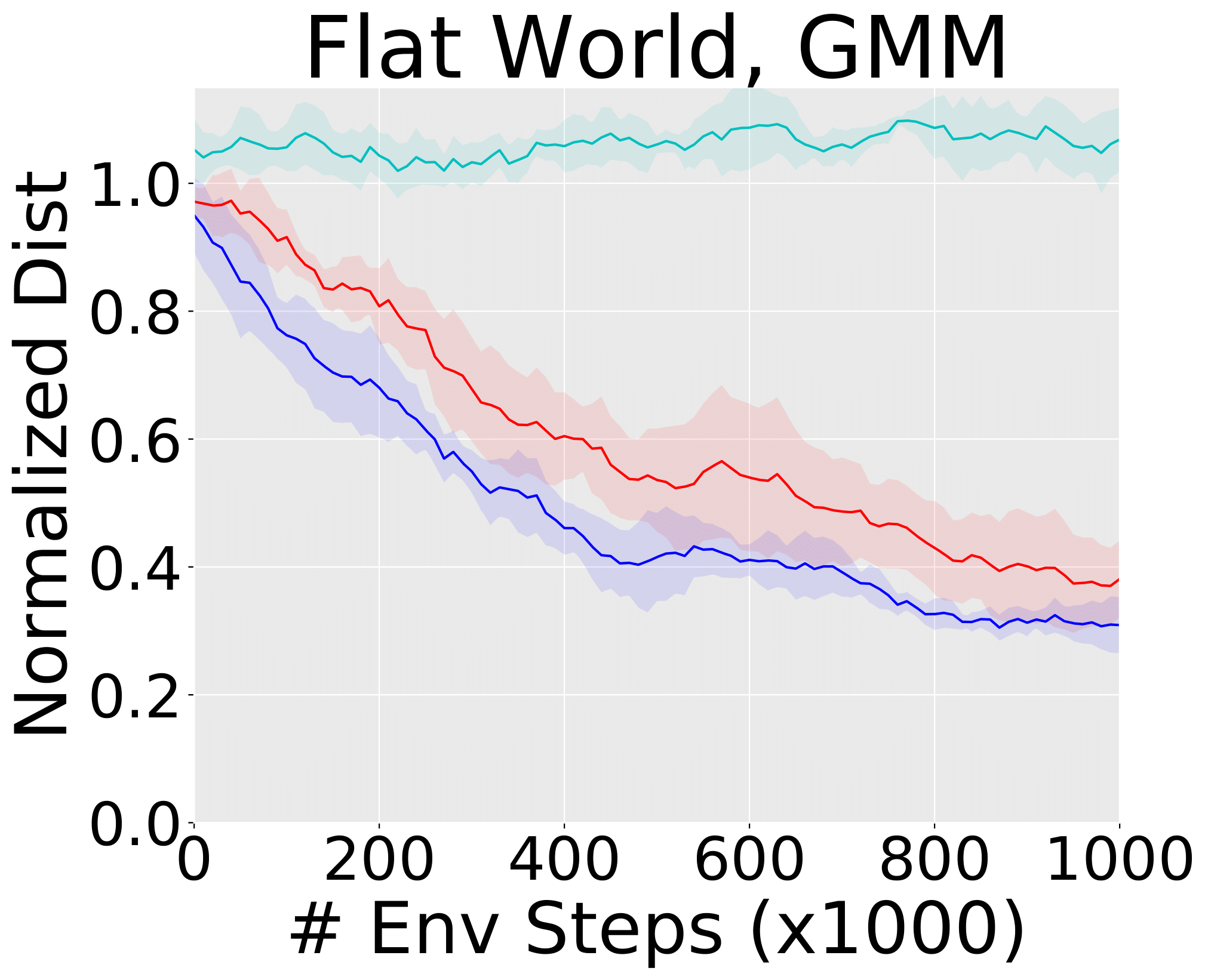}
    \hfill
    \vspace{0.1cm}
    \includegraphics[height=2.2cm]{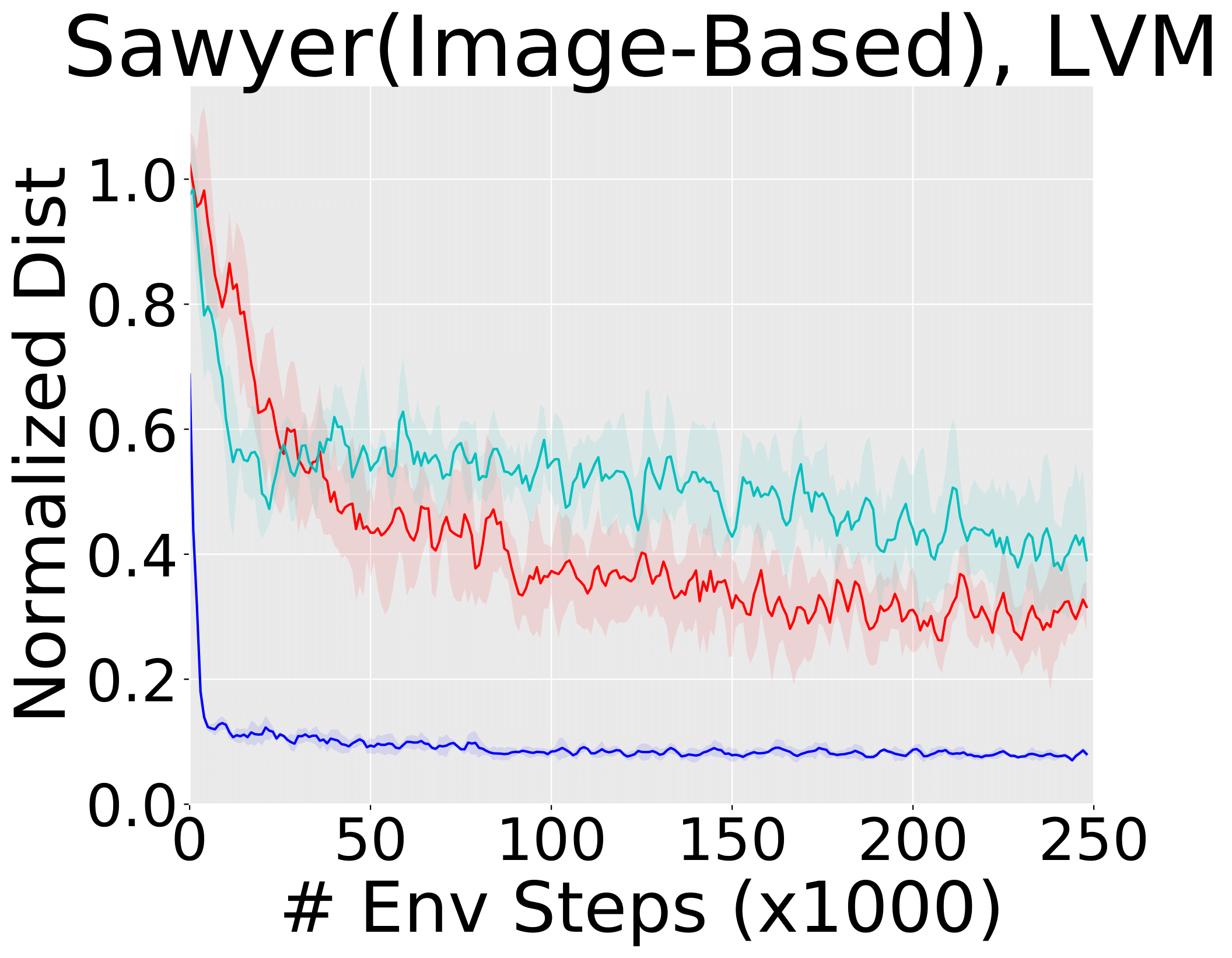}
    \includegraphics[width=0.70\linewidth]{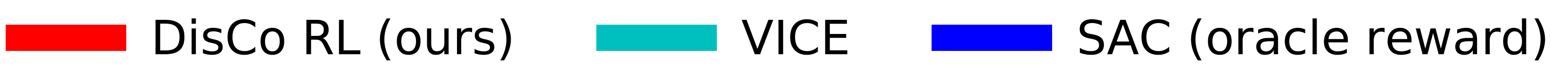}
    \hfill
    \caption{
    (Lower is better.)
    Learning curve showing normalized distance versus environment steps of various method.
    \METHOD uses a (left) Gaussian model, (middle) Gaussian mixture model, or (right) latent-variant model on their respective tasks. \METHOD with a learned goal distribution consistently outperforms VICE and obtains a final performance similar to using oracle rewards.
    }
    \vspace{-15pt}
    \label{fig:single-task-results}
\end{figure}

We see in \autoref{fig:single-task-results} that \METHOD often matches the performance of using an oracle reward and consistently outperforms VICE.
VICE often failed to learn, possibly because the method was developed using hundreds to thousands of examples, where as we only provided $30$ to $50$ examples.

\subsection{Conditional Distributions for Sub-Task Decomposition}
The next experiments study how \CMETHOD can automatically decompose a complex task into easier sub-tasks.
We design tasks that require reaching a desired final state $\bs_f$, but that can be decomposed into smaller sub-tasks.
The first task requires controlling the Sawyer robot to move $4$ blocks with randomly initialized positions into a box at a fixed location.
We design analogous tasks in the IKEA environment (with $3$ shelves and a pole) and Flat World environment (with $4$ objects).
All of these tasks can be split into sub-task that involving moving a single object at a time.

For each object $i=1, \dots, M$, we collect an example set $\dtask^i$.
As described in \Cref{sec:infer-conditional}, each set $\dtask^i$ contains $K=30$ to $50$ pairs of state $(\bs, \bs_f)$, in which object $i$ is in the same location as in the final desired state $\bs_f$, as shown in \Cref{fig:conditional-distribution-visualization}.
We fit a joint Gaussian to these pairs using \Cref{eq:learn-joint-distrib}.
During exploration and evaluation, we sample an initial state $\bs_0$ and final goal state $\bs_f$ uniformly from the set of possible states, and condition the policy on $\gdparam$ given by \Cref{eq:conditional-params}.
Because the tasks are randomly sampled, the specific task and goal distribution at evaluation are unseen during training, and so this setting tested whether \CMETHOD can generalize to new goal distributions.

For evaluation, we test how well \METHOD can solve long-horizon tasks by conditioning the policy on each \mbox{$p_g^i(\bs \mid \bs_f)$} sequentially for $H/M$ time steps.
We report the cumulative number of tasks that were solved, where each task is considered solved when the respective object is within a minimum distance of its target location specified by $\bs_f$.
For the IKEA environment, we also consider moving the pole to the correct location as a task.
We compare to VICE which trains separate classifiers and policies for each sub-task using the examples and also sequentially runs each policies for $H/M$ time steps.

\begin{figure}[t]
\centering
    \includegraphics[height=2.2cm]{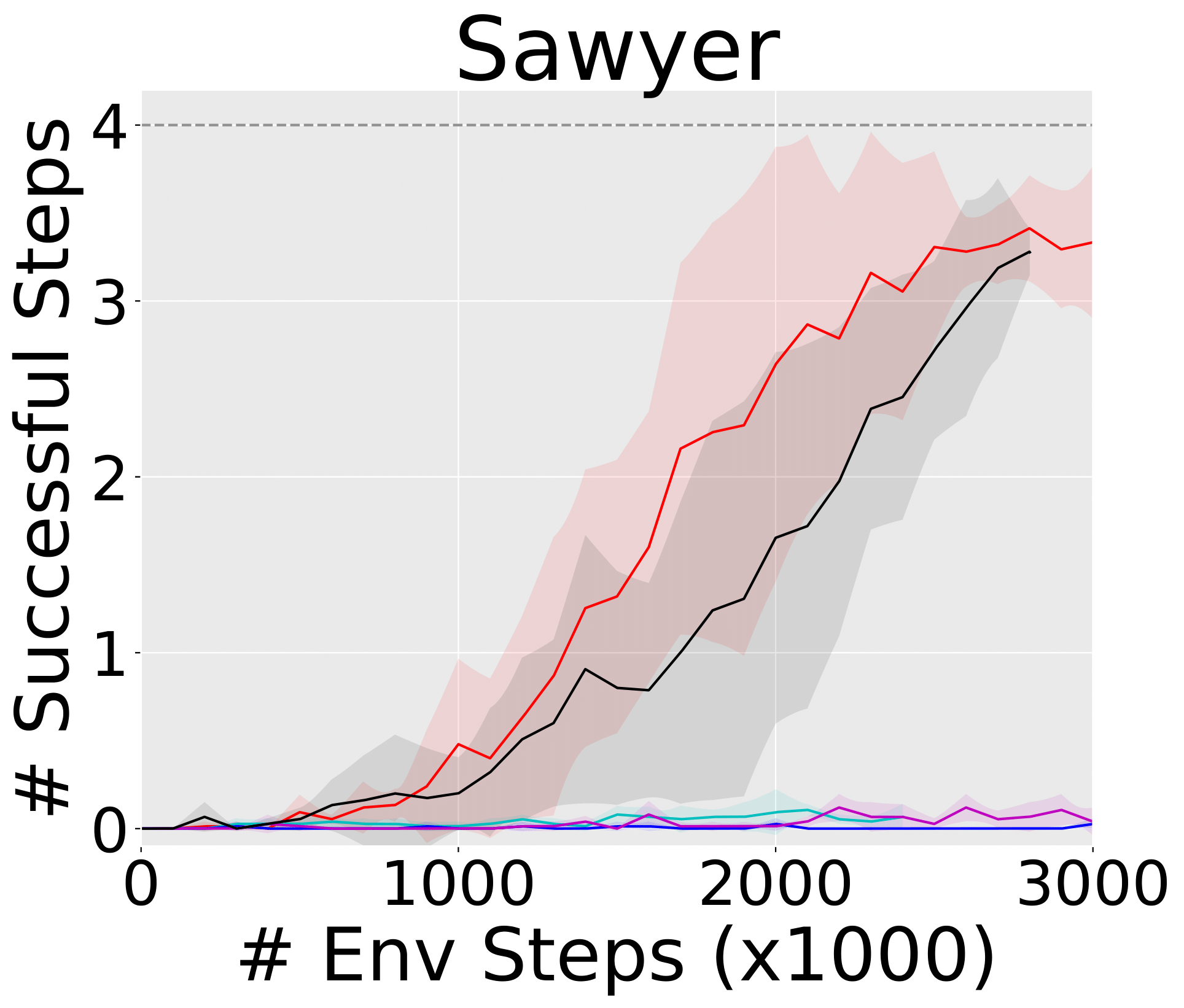}
    \hfill
    \includegraphics[height=2.2cm]{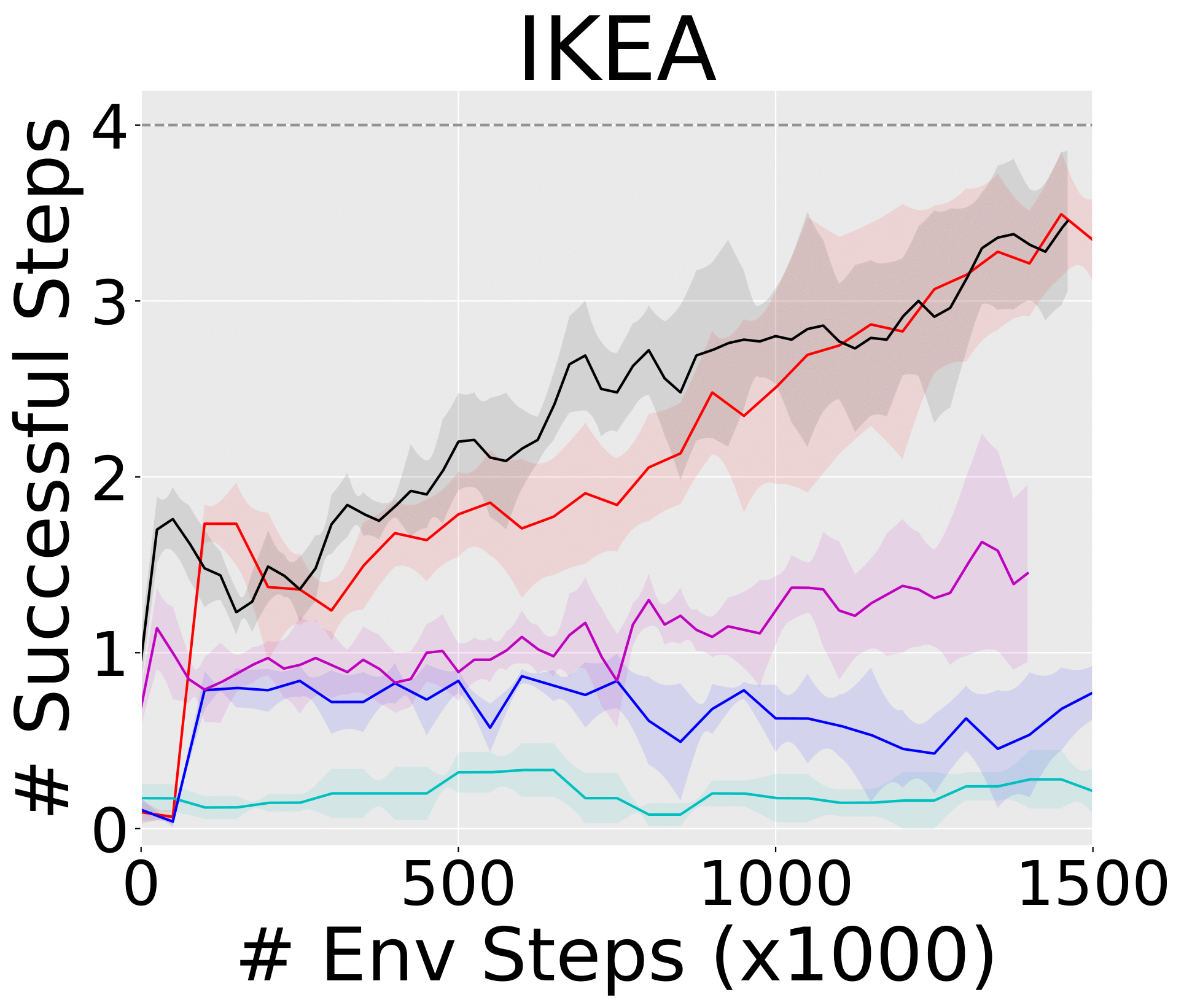}
    \hfill
    \vspace{0.1cm}
    \includegraphics[height=2.2cm]{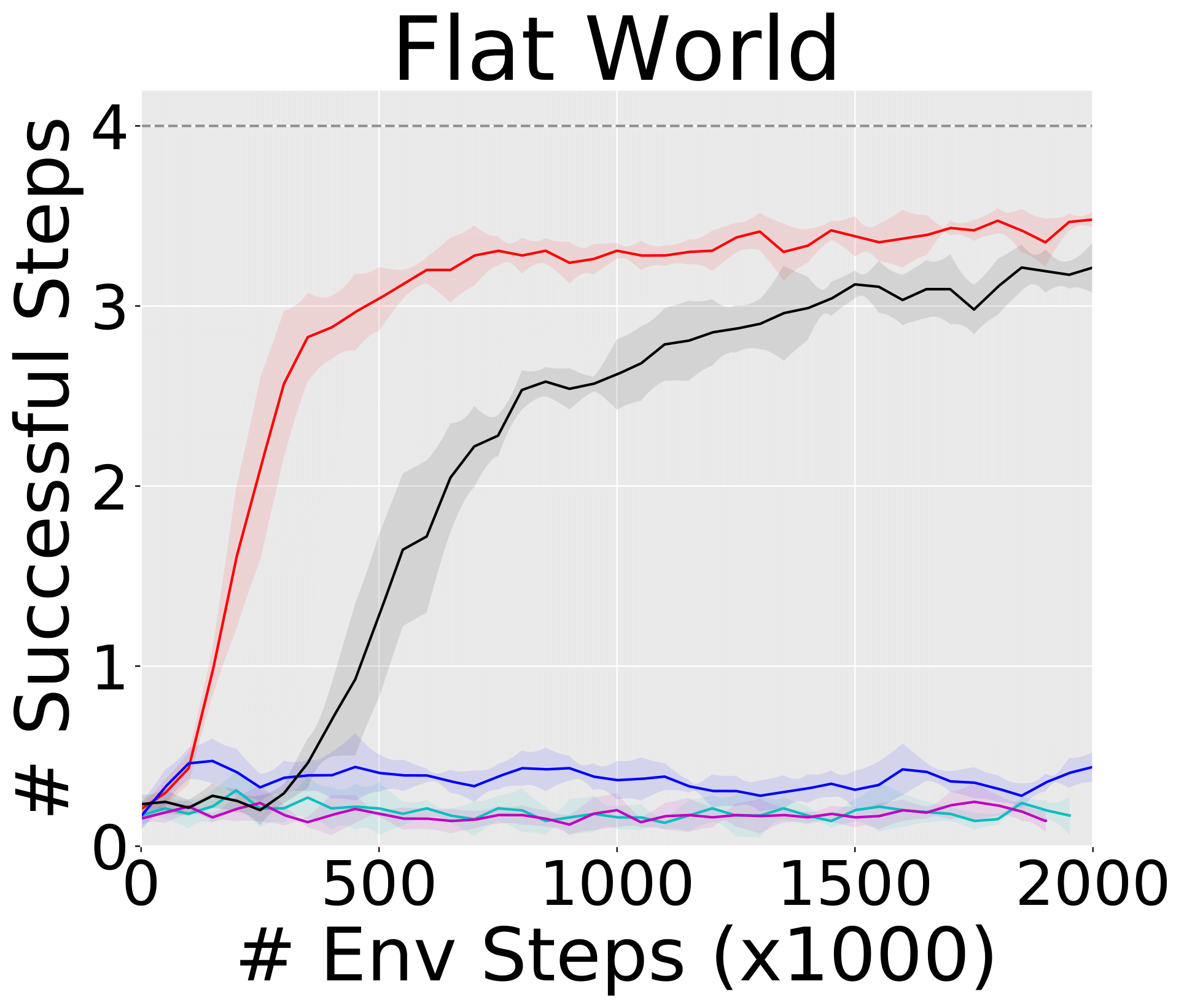}
    \includegraphics[width=\linewidth]{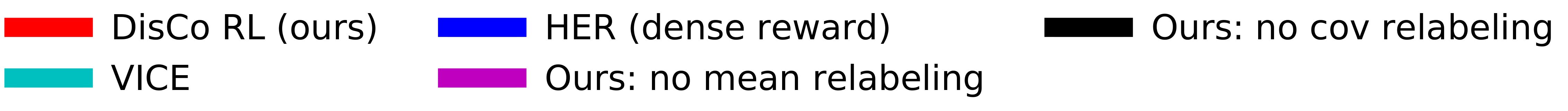}
        
    \caption{(Higher is better.) Learning curves showing the number of cumulative tasks completed versus environment steps for the Sawyer (left), IKEA (middle), and Flat World (right) tasks. We see that \METHOD significantly outperforms HER and VICE, and that relabeling the mean and covariance is important.
    }
    \label{fig:long-horizon-results}
    \vspace{-0.50cm}
\end{figure}

One can train a goal-conditioned policy to reach this final state $\bs_f$ directly, and so we compare to hindsight experience replay (\textbf{HER})~\citep{andrychowicz2017her}, which attempts to directly reach the final state for $H$ time steps and learns using an oracle dense reward.
Since HER does not decompose the task into subtasks, we expect that the learning will be significantly slower since it does not exploit access to the tuples that give examples of a subtask being completed.
We see in \autoref{fig:long-horizon-results} that \CMETHOD significantly outperforms VICE and HER and that \CMETHOD successfully generalizes to new goal distributions.

\paragraph{Ablations}
Lastly, we include ablations that test the importance of relabeling the mean and covariance parameters during training. 
We see in \autoref{fig:long-horizon-results} that relabeling both parameters, and particularly the mean, accelerates learning.

\section{Discussion}\label{sec:conclusion}
We presented \METHOD, a method for learning general-purpose policies specified using a goal distribution.
Our experiments show that \METHODADJ policies can solve a variety of tasks using goal distributions inferred from data, and can accomplish tasks specified by goal distributions that were not seen during training.
An exciting direction for future work would be to interleave \METHOD and distribution learning, for example by using newly acquired data to update the learned VAE or by backpropagating the \METHOD loss into the VAE learning loss.
Another promising direction would be to study goal-distribution directed exploration, in which an agent can explore along certain distribution of states, analogous to work on goal-directed exploration~\citep{koenig1996effect,ecoffet2019go,ren2019exploration,pong2019skewfit}.

\\\\
\noindent \textbf{Acknowledgements}
This research was supported by the Office of Naval Research, the National Science Foundation through IIS-1651843, and Berkeley DeepDrive.
\newpage



{ \small
\bibliographystyle{IEEEtran}
\bibliography{references}
}

\end{document}